%% file: main.tex
\documentclass{article}
\usepackage{ijcai21}

\usepackage{times}
\usepackage{soul}
\usepackage{url}
\usepackage[hidelinks]{hyperref}
\usepackage[utf8]{inputenc}
\usepackage[small]{caption}
\usepackage{graphicx}
\usepackage{amsmath}
\usepackage{amsthm}
\usepackage{booktabs}
\usepackage{algorithm}
\usepackage{algorithmic}
\title{Physics-aware Spatiotemporal Modules with Auxiliary Tasks for Meta-Learning}

\author{
Sungyong Seo
\and
Chuizheng Meng\and
Sirisha Rambhatla\And
Yan Liu
\affiliations
University of Southern California, USA\\
\emails
\{sungyons, chuizhem, sirishar, yanliu.cs\}@usc.edu
}
\usepackage{wrapfig}
\usepackage{multirow}
\usepackage{amsmath,amsfonts,amssymb}       % blackboard math symbols
\usepackage{subfigure}

\usepackage{color}
\usepackage{xcolor}

\newcommand{\ours}{{PiMetaL}}

\definecolor{OliveGreen}{rgb}{0,0.6,0}

\input{math_commands}

\begin{document}

\maketitle

% \normalsize
\input{abstract}

\input{introduction}

\input{problem}

\input{sdm-generalization}

\input{experiment}

\input{related-work}

\input{conclusion}

\section*{Acknowledgements}
This work is supported in part by NSF Research Grant IIS-1254206 and MINERVA grant N00014-17-1-2281, granted to co-author Yan Liu in her academic role at the University of Southern California.
The views and conclusions are those of the authors and should not be interpreted as representing the official policies of the funding agency, or the U.S. Government.
% Last but not least, we appreciate anonymous reviewers for your thorough comments and suggestions.

%% The file named.bst is a bibliography style file for BibTeX 0.99c
% \newpage
\small
% \footnotesize
\bibliographystyle{named}
\bibliography{references}

% \newpage
% \normalsize
% \appendix
% \input{appendix}

\end{document}

%% file: math_commands.tex
\usepackage{amsmath,amsfonts,bm}

\def\eqref#1{equation~\ref{#1}}
\def\1{\bm{1}}

\def\ve{{\bm{e}}}

\def\vg{{\bm{g}}}

\def\vu{{\bm{u}}}
\def\vv{{\bm{v}}}

\def\vx{{\bm{x}}}

\DeclareMathAlphabet{\mathsfit}{\encodingdefault}{\sfdefault}{m}{sl}
\SetMathAlphabet{\mathsfit}{bold}{\encodingdefault}{\sfdefault}{bx}{n}

\def\gD{{\mathcal{D}}}

\def\gG{{\mathcal{G}}}

\def\sK{{\mathbb{K}}}

\def\sV{{\mathbb{V}}}

%% file: abstract.tex
\begin{abstract}
Modeling the dynamics of real-world physical systems is critical for spatiotemporal prediction tasks, but challenging when data is limited. The scarcity of real-world data and the difficulty in reproducing the data distribution hinder directly applying meta-learning techniques. Although the knowledge of governing partial differential equations (PDE) of data can be helpful for the fast adaptation to few observations, it is mostly infeasible to exactly find the equation for observations in real-world physical systems. In this work, we propose a framework, physics-aware meta-learning with auxiliary tasks, whose spatial modules incorporate PDE-independent knowledge and temporal modules utilize the generalized features from the spatial modules to be adapted to the limited data, respectively. The framework is inspired by a local conservation law expressed mathematically as a continuity equation and does not require the exact form of governing equation to model the spatiotemporal observations. The proposed method mitigates the need for a large number of real-world tasks for meta-learning by leveraging spatial information in simulated data to meta-initialize the spatial modules. We apply the proposed framework to both synthetic and real-world spatiotemporal prediction tasks and demonstrate its superior performance with limited observations.
\end{abstract}

%% file: introduction.tex
\section{Introduction}
Deep learning has recently shown promise to play a major role in devising new solutions to applications with natural phenomena, such as climate change~\cite{drgona2019stripping}, ocean dynamics~\cite{cosne2019coupling}, air quality~\cite{soh2018adaptive}, and so on. 
Deep learning techniques inherently require a large amount of data for effective representation learning, so their performance is significantly degraded when there are only a limited number of observations.
However, in many tasks in the real-world physical systems we only have access to a limited amount of data.
One example is air quality monitoring~\cite{aqi2017}, in which the sensors are irregularly distributed over the space -- many sensors are located in urban areas whereas there are much fewer sensors in vast rural areas. 
Another example is extreme weather modeling and forecasting, i.e., temporally short events (e.g., tropical cyclones~\cite{racah2017extremeweather}) without sufficient observations over time.
% The situation is further exacerbated for sensor-based dataset (spatially low resolution) or extreme weather events (temporally short).
% It is even more limited for sensor-based dataset since a target (e.g., temperature) is only observed at where a sensor is located (spatially low resolution).
Moreover, inevitable missing values from sensors~\cite{tang2019joint} further reduce the number of operating sensors and shorten the length of fully-observed sequences.
Thus, achieving robust performance from a few spatiotemporal observations in physical systems remains an essential but challenging problem.

Learning on a limited amount of data from physical systems can be considered as a few shot learning.
While many meta-learning techniques~\cite{ravi2016optimization,finn2017model,munkhdalai2017meta,duan2017one,snell2017prototypical,mishra2018a} have been developed to handle this few shot learning setting, there are still some challenges for the existing meta-learning methods to be applied in modeling natural phenomena.
First, it is not easy to find a set of similar meta-tasks, which provide shareable latent representations needed to understand targeted observations.
For instance, while image-related tasks (object detection~\cite{he2017mask}) can take advantage of an image-feature extractor pre-trained by a large set of images~\cite{deng2009imagenet} and well-designed architecture~\cite{he2016deep}, there is no such large data corpus that is widely applicable for understanding natural phenomena.
Second, unlike computer vision or natural language processing tasks where a common object (images or words) is clearly defined, it is not straightforward to find analogous objects in the spatiotemporal data.
Finally, exact equations behind natural phenomena are usually unknown, causing the difficulty in reproducing the similar dataset via simulation.
Although there have been some works~\cite{de2018deep,greydanus2019hamiltonian} improving data efficiency via explicitly incorporating PDEs as neural network layers when modeling spatiotemporal dynamics, it is hard to generalize for modeling different or unknown dynamics, which is ubiquitous in real-world scenario.

In this work, we propose physics-aware modules designed for meta-learning to tackle the few shot learning challenges in physical observations.
One of fundamental equations in physics describing the transport of physical quantity over space and time is a continuity equation:
\begin{align}
    \frac{\partial\rho}{\partial t}+\nabla\cdot J=\sigma,\label{eq:continuity-eqn}
\end{align}
where $\rho$ is the amount of the target quantity ($u$) per unit volume, $J$ is the flux of the quantity, and $\sigma$ is a source or sink, respectively.
This fundamental equation can be used to derive more specific transport equations such as the convection-diffusion equation, Navier-Stokes equations, and Boltzmann transport equation.
Thus, the continuity equation is a starting point to model spatiotemporal (conservative) observations which are accessible from sensors.
Based on the form of $\rho$ and $J$ in terms of $u$, Eq.~\ref{eq:continuity-eqn} can be generalized as:
\begin{align}
    \frac{\partial u}{\partial t} = F(\nabla u,\nabla^2u,\dots),\label{eq:pde-general}
    % {\partial u}/{\partial t} = F(u_x,u_y,u_{xx},u_{yy},\dots),\label{eq:pde-general}
\end{align}
where the function $F(\cdot)$ describes how the target $u$ is changed over time from its spatial derivatives.
Inspired by the form of Eq.~\ref{eq:pde-general}, we propose two modules: spatial derivative modules (SDM) and time derivative modules (TDM).
Since the spatial derivatives such as $\nabla, \nabla\cdot,$ and $\nabla^2$ are commonly used across different PDEs, the spatial modules are PDE-independent and they can be meta-initialized from synthetic data.
Then, the PDE-specific temporal module is trained to learn the unknown function $F(\cdot)$ from few observations in the real-world physical systems.

This approach can effectively leverage a large amount of simulated data to train the spatial modules as the modules are PDE-independent and thus mitigating the need for a large amount of real-world tasks to extract shareable features.
In addition, since the spatial modules are universally used in physics equations, the representations from the modules can be conveniently integrated with data-driven models for modeling natural phenomena.
% In addition, since the spatial modules are universally used in physics equations and explicit forms of equations are not required for training the modules, the modular method can conveniently integrate meta-learning for natural phenomena.
Based on the modularized PDEs, we introduce a novel approach that marries physics knowledge in spatiotemporal prediction tasks with meta-learning by providing shareable modules across spatiotemporal observations in the real-world.

Our contributions are summarized below:
\begin{itemize}
    \item \textbf{Modularized PDEs and auxiliary tasks:} Inspired by forms of PDEs in physics, we decompose PDEs into shareable (spatial) and adaptation (temporal) parts.
    The shareable one is PDE-independent and specified by auxiliary tasks: \textit{supervision of spatial derivatives}.
    % Spatial derivatives can be approximated by PDE-independent modules with corresponding auxiliary tasks from simulated data, which can be from a different data distribution and is easy to generate.
    \item \textbf{Physics-aware meta-learning:} We provide a framework for physics-aware meta-learning, which consists of PDE-independent/-specific modules. 
    %The framework shows how physics-related regression tasks can be adapted to meta-learning settings.
    The framework is flexible to be applied to the modeling of different and unknown dynamics.
    % The framework enables to bring physics-related regression tasks into meta-learning settings.
    \item \textbf{Utilizing synthetic data for shareable modules:} We extract shareable parameters in the spatial modules from synthetic data, which can be easily generated.
    % which are helpful for real-world spatiotemporal prediction tasks.
    % Spatial derivatives can be approximated by PDE-independent modules with corresponding auxiliary tasks from simulated data, which can be from a different data distribution and is easy to generate.
    % We evaluate our framework on synthetic and three sensor-based datasets (few observations) governed by unknown physics rules for spatiotemporal prediction tasks.
\end{itemize}

% Our main contributions are:
% \begin{itemize}
%     \item Modular meta-learning: task-independent module, which learns finite difference coefficients to approximate spatial derivatives from function values.
%     \item Sparsely available spatial data: 
%     \item 
% \end{itemize}

% paradox

% A module is task-independent, thus, a limited number of data points in a main task doesn't matter.
% It can be trained by a large number of task-independent data.

%% file: problem.tex
\section{Modularized PDEs and Meta-Learning}

In this section, we describe how the physics equations for conserved quantities are decomposable into two parts and how the meta-learning approach tackles the task by utilizing synthetic data when the data are limited.

\subsection{Decomposability of Variants of a Continuity Equation}\label{subsec:decompose}
In physics, a continuity equation (Eq.~\ref{eq:continuity-eqn}) describes how a locally conserved quantity such as temperature, fluid density, heat, and energy is transported across space and time.
This equation underlies many specific equations such as the convection-diffusion equation and Navier-Stokes equations:
\begin{align*}
    & \dot{u} = \nabla\cdot(D\nabla u)-\nabla\cdot(\vv u) + R,\\
    & \dot{\vu} = -(\vu\cdot\nabla)\vu + \nu\nabla^2\vu - \nabla\omega + \vg.
\end{align*}
where the scalar $u$ and vector field $\vu$ are the variables of interest (e.g., temperature, flow velocity, etc.). 
A dot over a variable is time derivative.
The common feature in these equations is that the forms of equations can be digested as \cite{bar2019learning}:
\begin{align}
    \dot{u} = F(u_x,u_y,u_{xx},u_{yy},\dots),\label{eq:pde-digest}
\end{align}
where the right-hand side denotes a function of spatial derivatives.
As the time derivative can be seen as a Euler discretization~\cite{chen2018neural}, it is notable that the next state is a function of the current state and spatial derivatives.
Thus, knowing spatial derivatives at time $t$ is a key step for spatiotemporal prediction at time $t+1$ for locally conserved quantities.
According to Eq.~\ref{eq:pde-digest}, the spatial derivatives are universally used in variants of Eq.~\ref{eq:continuity-eqn} and only the updating function $F(\cdot)$ is specifically defined for a particular equation.
This property implies that PDEs for physical quantities can be decomposable into two modules: spatial and temporal derivative modules.

%%%%
\subsection{Spatial Derivative Modules: PDE-independent Modules}\label{sec:fdm}
Finite difference method (FDM) is widely used to discretize a $d$-order derivative as a linear combination of function values on a $n$-point stencil.
\begin{align}
    \frac{\partial^d u}{\partial x^d} \approx \sum_{i=1}^{n}\alpha_i u(x_i),\label{eq:fdm}
\end{align}
where $n>d$.
According to FDM, it is independent for a form of PDE to compute spatial derivatives, which are input components of $F(\cdot)$ in Eq.~\ref{eq:pde-digest}.
Thus, we can modularize spatial derivatives as PDE-independent modules.
The modules that can be learnable as a data-driven manner to infer the coefficients ($\alpha_i$) have been proposed recently~\cite{bar2019learning}.
The data-driven coefficients are particularly useful when the discretization in the $n$-point stencil is irregular and low-resolution where the fixed coefficients cause substantial numerical errors.

\subsection{Time Derivative Module: PDE-specific Module}
Once $d$-order derivatives are modularized by learnable parameters, the approximated spatial derivatives from the spatial modules are fed into an additional module to learn the function $F(\cdot)$ in Eq.~\ref{eq:pde-digest}.
This module is PDE-specific as the function $F$ describes how the spatiotemporal observations change.
Since the exact form of a ground truth PDE is not given, the time derivative module is data-driven and will be adapted to observations instead.

\subsection{Meta-Learning with PDE-independent/-specific Modules}
Recently,~\cite{raghu2020Rapid} investigate the effectiveness of model agnostic meta-learning (MAML,~\cite{finn2017model}) and it is found that the outer loop of MAML is more likely to learn parameters for reusable features rather than rapid adaptation.
The finding that feature reuse is the predominant reason for efficient learning of MAML allows us to use additional information which is beneficial for learning better representations. 
Previously, the objective in meta-training has been considered to be matched with one in meta-test as the purpose of meta-learning is to learn good initial parameters applicable across similar tasks (e.g., image classification to image classification). 
We are now able to incorporate auxiliary tasks under a meta-learning setting to reinforce reusable features for a main task.
As described in Sec.~\ref{subsec:decompose}, the spatial modules are reusable across different observations, and thus, we can meta-initialize the spatial modules first with spatial derivatives provided by synthetic datasets. 
Then, we can integrate the spatial modules with the task-specific temporal module during meta-test to help adaptation of TDM on few observations.
Since the spatial modules are trained by readily available synthetic datasets, a large number of real-world datasets for meta-training is not required.

\section{Physics-aware Meta-Learning with Auxiliary Tasks}\label{sec:mml-aux}

\begin{figure}[t]
    \centering
\includegraphics[width=0.95\linewidth]{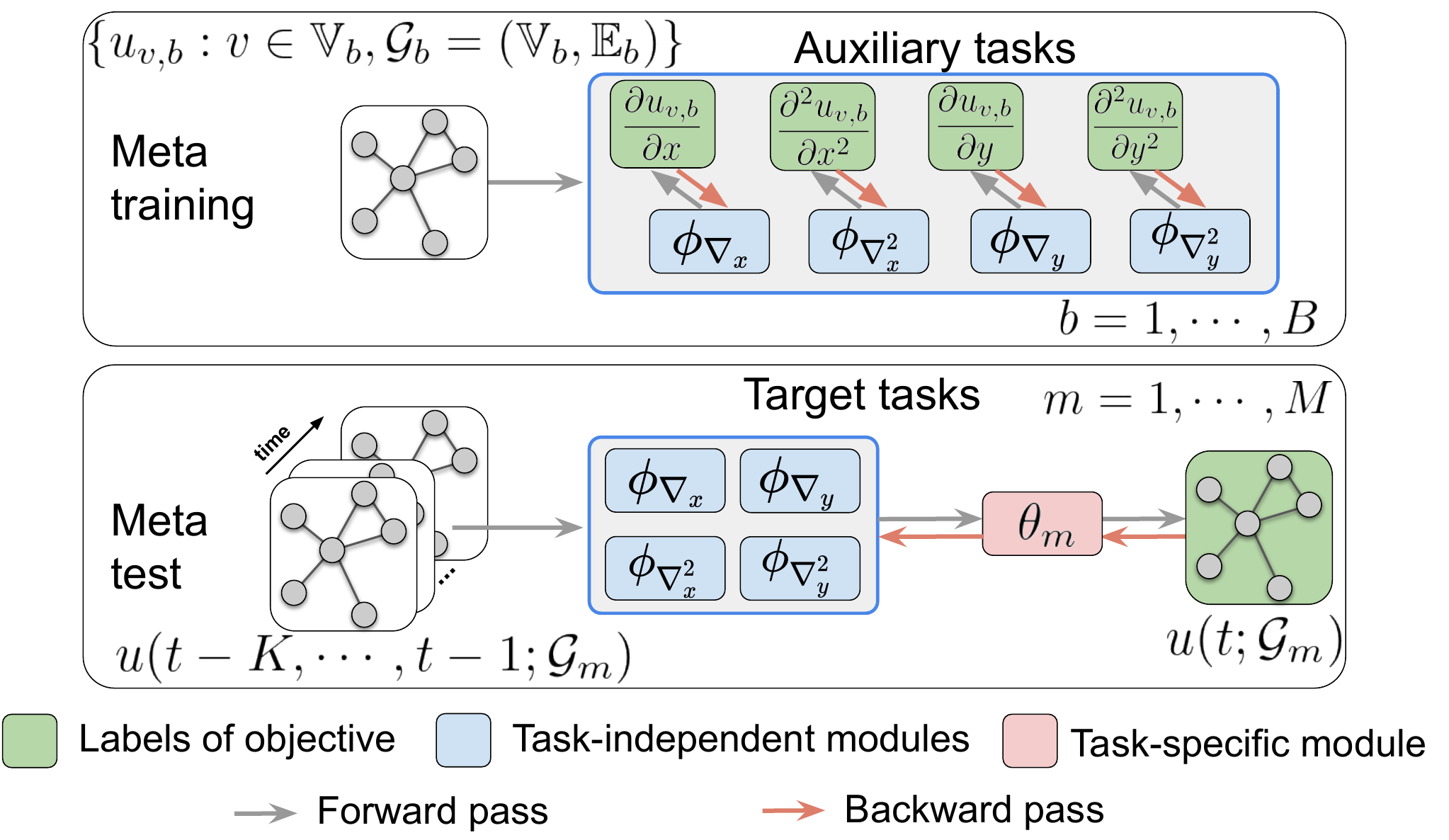}
    \caption{Overview of the physics-aware meta-learning (\ours).}
    \label{fig:pimeta}
    \vspace{-2mm}
\end{figure}

In this section, we develop a physics-aware meta-learning framework for the modularized PDEs.
Fig.~\ref{fig:pimeta} describes the proposed framework and its computational process.

%%%%%%
%\subsection{Structured modular meta-learning}

%%%% old
% In this section, we provide a form of structured modular meta-learning~\cite{alet2018modular,alet2019neural} and extend it to a generalized framework: modular meta-learning with auxiliary objective (\mmla), which can be reduced to the structured modular meta-learning as a special case.
% Suppose that $\gD=(\gD_1,\dots,\gD_M)$ is a set of meta-learning dataset where $M$ is the number of main tasks.
% Note that the term \textit{main tasks} indicate a set of tasks we eventually want a model to perform well with fewer data and it will be distinguishable to an auxiliary task described in Section~\ref{sec:mml-aux}.
%%%%
%%%%%
% \subsection{Structured modular meta-learning}
% The framework of structured modular meta-learning is introduced in~\cite{alet2018modular,alet2019neural}. 

%%%%
\subsection{Spatial Derivative Module}\label{subsec:sdm}
% \vspace{-1em}

As we focus on the modeling and prediction of sensor-based observations, where the available data points are inherently on a spatially sparse irregular grid, we use graph networks for each module $\phi_k$ to learn the finite difference coefficients.
Given a graph $\gG=(\sV,\mathbb{E})$ where $\sV=\{1,\dots,N\}$ and $\mathbb{E}=\{(i,j): i,j\in\sV\}$, a node $i$ denotes a physical location $\vx_i=(x_i,y_i)$ where a function value $u_i=u(x_i,y_i)$ is observed.
% The observation at node $i$ is $u(x_i,y_i)$ or $u_i$.
Then, the graph signals with positional relative displacement as edge features are fed into the spatial modules to approximate spatial derivatives by Alg.~\ref{alg:sdm}.
The coefficients $(a_i,b_{(i,j)})$ on each node $i$ and edge $(i,j)$ are output of $\phi$ and they are linearly combined with the function values $u_i$ and $u_j$.
$\sK$ denotes a set of finite difference operators.
For example, if we set $\sK=\{\nabla_x,\nabla_y,\nabla^2_{x},\nabla^2_{y}\}$, we have 4 modules which approximate first/second order of spatial derivatives in 2-dimension, respectively.

\begin{algorithm}[t]
   \caption{Spatial derivative module (SDM)}
   \label{alg:sdm}
   \textbf{Input}: Graph signals $u_i$ and edge features $\ve_{ij}=\vx_j-\vx_i$ on $\gG$ where $\vx_i$ is a coordinate of node $i$. \\
%   A set of module attributes $\sK$\\
   \textbf{Output}: Spatial derivatives $\{\hat{u}_{k,i}\ \vert \ i\in\sV\ \text{and}\ k\in\sK \}$ where $\sK=\{\nabla_x,\nabla_y,\nabla^2_{x},\nabla^2_{y}\}$.
   \begin{flushleft}
    \textbf{Require}: Spatial derivative module $\{\phi_k \vert k\in\sK\}$.
   \end{flushleft}
    \begin{algorithmic}[1]
        \FOR{$k\in\sK$}
        \STATE $\left\lbrace a_{k,i},b_{k,(i,j)}\ \vert\ (i,j)\in\mathbb{E}, k\in\sK\right\rbrace$ = $\phi_k(\left\lbrace u \right\rbrace, \left\lbrace \ve \right\rbrace, \mathcal{G})$
        \FOR{$i\in\sV$}
        % \STATE $\hat{u}_{k,i} = \sum_{(i,j)\in\mathbb{E}} a_{k,(i,j)}(u_i - b_{k,(i,j)}\cdot u_j)$
        \STATE $\hat{u}_{k,i} = a_{k,i}u_i + \sum_{(j,i)\in\mathbb{E}} b_{k,(j,i)}u_j$
        \ENDFOR
        % \STATE Update $\phi_k\leftarrow\phi_k-\alpha\sum_{i\in\mathbb{V}}$
        \ENDFOR
    \end{algorithmic}
\end{algorithm}

%%%%

\begin{algorithm}[t]
   \caption{Time derivative module (TDM)}
   \label{alg:tdm}
   \begin{flushleft}
   \textbf{Input}: Graph signals $u$ and approximated spatial derivatives $\hat{u}_{k}$ where $k\in\sK$ on $\gG$. Time interval $\Delta t$\\
   \textbf{Output}: Prediction of signals at next time step $\hat{u}(t)$\\
   \textbf{Require}: Time derivative module.
   \end{flushleft}
\begin{algorithmic}[1]
    \STATE $\hat{u}_t$ = \text{TDM}$\left(\left\lbrace u_i, \hat{u}_{k,i}\ \vert\ i\in\sV\ \text{and}\ k\in\sK \right\rbrace\right)$
    \STATE $\hat{u}(t)$ = $u(t-1) + \hat{u}_{t-1}\cdot \Delta t$
\end{algorithmic}
\end{algorithm}

\subsection{Time Derivative Module}\label{subsec:tdm}
Once spatial derivatives are approximated, another learnable module is required to combine them for a target task.
The form of line 2 in Alg.~\ref{alg:tdm} comes from Eq.~\ref{eq:pde-digest} and TDM is adapted to learn the unknown function $F(\cdot)$ in the equation.
% The exact form of $F$ for the target task is not restricted.
As our target task is the regression of graph signals, we use a recurrent graph network for TDM.

%%%%%%

\subsection{Meta-Learning with Auxiliary Objective}

As discussed in Sec.~\ref{subsec:decompose}, it is important to know spatial derivatives at time $t$ to predict next signals at $t+1$ for locally conserved physical quantities.
However, it is impractical to access the spatial derivatives in the sensor-based observations as they are highly discretized over space.
% We propose a physics-aware meta-learning framework to meta-initialize spatial modules by leveraging synthetic dataset with auxiliary tasks to provide reusable features for the main tasks: prediction spatiotemporal observations in the real-world.

The meta-initialization with the auxiliary tasks from synthetic datasets is particularly important to address the challenge.
First, the spatial modules can be universal feature extractors for modeling observations following unknown physics-based PDEs. 
Unlike other domains such as computer vision, it has been considered that there is no particular shareable architecture for learning spatiotemporal dynamics from physical systems.
We propose that the PDE-independent spatial modules can be applicable as feature extractors across different dynamics as long as the dynamics follow a local form of conservation laws.
Second, we can utilize synthetic data to meta-train the spatial modules as they are PDE-agnostic.
This property allows us to utilize a large amount of synthetic datasets which are readily generated by numerical methods regardless of the exact form of PDE for target observations.
Finally, we can provide a stronger inductive bias which is beneficial for modeling real-world observations but not available in the observations explicitly.

\begin{figure}[t]
    \centering
    \subfigure{\includegraphics[width=0.47\linewidth]{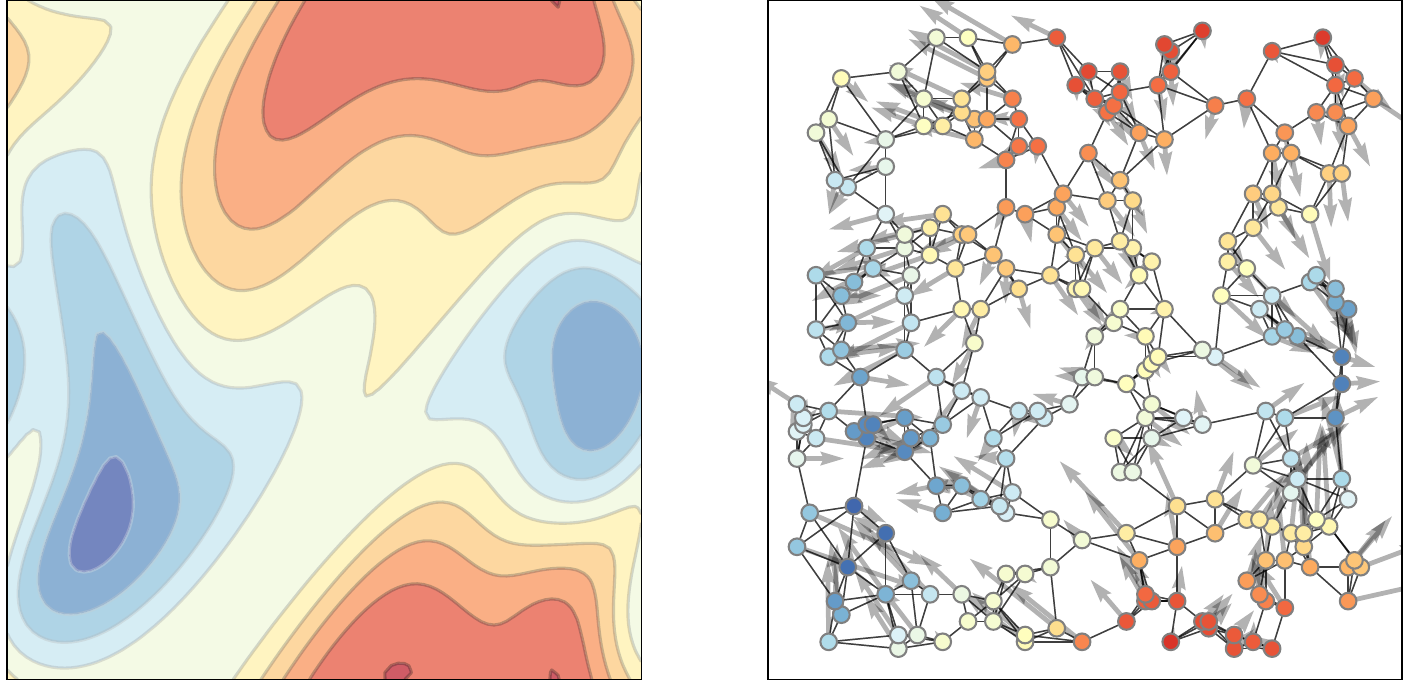}}
    \hfill
    \subfigure{\includegraphics[width=0.47\linewidth]{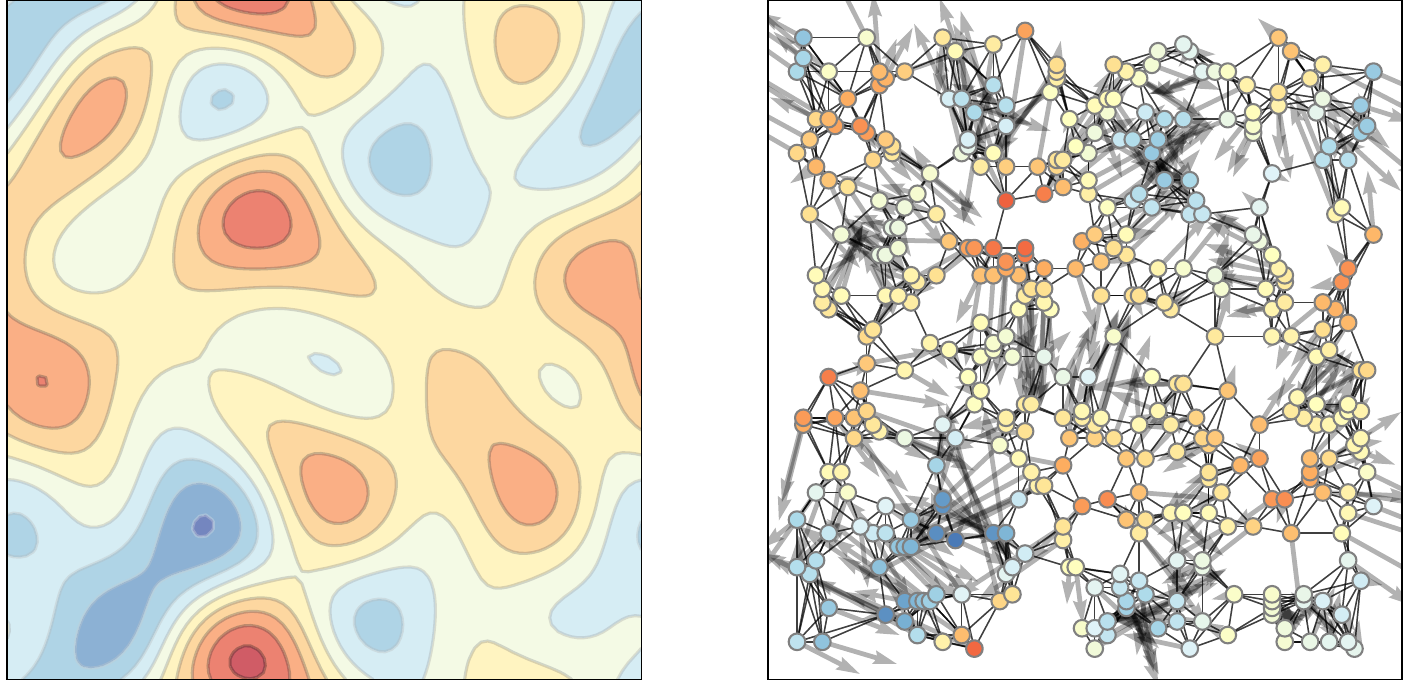}}
    % \hfill
    % \subfigure{\includegraphics[width=0.47\linewidth]{fig/samples-aux-data2.pdf}}
    \caption{Examples of generated spatial function values and graph signals. Node and edge features (function value and relative displacement, respectively) are used to approximate spatial derivatives (arrows). We can adjust the number of nodes (spatial resolution), the number of edges (discretization), and the degree of fluctuation (scale of derivatives) to differentiate meta-train tasks.}
    \label{fig:aux}
    \vspace{-2mm}
\end{figure}

Alg.~\ref{alg:mml-aux-train} describes how the spatial modules are meta-initialized by MAML under the supervision of $K$ different spatial derivatives.
First, we generate values and spatial derivatives on a 2D regular grid from an analytical function.
Then, we sample a finite number of points from the regular grid to represent discretized nodes and build a graph from the sampled nodes.
Each graph signal and its discretization becomes input feature of a meta-train task and corresponding spatial derivatives are the auxiliary task labels.
Fig.~\ref{fig:aux} visualizes graph signals and spatial derivatives for meta-initialization.

\begin{algorithm}[t]
  \caption{Meta-initialization with auxiliary tasks: Supervision of spatial derivatives}
   \label{alg:mml-aux-train}
   \begin{flushleft}
    \textbf{Input}: A set of meta-train task datasets $\gD=\{\gD_1,\dots,\gD_B\}$ where $\gD_b=(\gD^{tr}_b,\gD^{te}_b)$. $\gD_b=\{(u^b_i,\ve^b_{ij}, y^{(a_1,b)}_i,\dots,y^{(a_K,b)}_i) : i\in\sV_b, (i,j)\in\mathbb{E}_b\}$ where $y^{(a_k,\cdot)}_i$ is an $k$-th auxiliary task label for the $i$-th node.
    % , given node/edge feature $u^b$ and $\ve^b$, respectively. 
    Learning rate $\alpha$ and $\beta$.\\
   \textbf{Output}: Meta-initialized spatial modules $\Phi$.
   \end{flushleft}
   \begin{algorithmic}[1]
   \STATE Initialize auxiliary modules $\Phi=(\phi_1,\dots,\phi_K)$
   \WHILE{not converged}
%   \STATE Sample a task $\gT_i$ from $\gT$
   \FOR{$\gD_b$ in $\gD$}
   \STATE $\Phi'_b=\Phi - \alpha\nabla_{\Phi}\sum_{k=1}^K\mathcal{L}^{aux}_{k}(\gD^{tr}_b;\phi_k)$
   \ENDFOR
   \STATE $\Phi\leftarrow\Phi-\beta\nabla_{\Phi}\sum_{b=1}^{B}\sum_{k=1}^K\mathcal{L}^{aux}_{k}(\gD^{te}_b;\phi'_{b,k})$
   \ENDWHILE
\end{algorithmic}
\end{algorithm}

Once the spatial modules are initialized throughout meta-training, we reuse the modules for meta-test where the temporal module (the head of the network) are adapted on few observations from real-world sensors (Alg.~\ref{alg:mml-aux-test}).
Although the standard MAML updates the body of the network (the spatial modules) as well, we only adapt the head layer ($\theta$) as like almost-no-inner-loop method in~\cite{raghu2020Rapid}.
The task at test time is graph signal prediction and the temporal modules ($\theta$) are adapted by a regression loss function $\mathcal{L}=\sum_{t=1}^{T}||u(t)-\hat{u}(t)||^2$ on length $T$ sequence ($\gD^{tr}_m$) and evaluated on held-out ($t>T$) sequence ($\gD^{te}_m$) with the adapted parameters.
% Inspired by~\cite{raghu2019rapid}, the head layer ($\theta$) is adapted.

\begin{algorithm}[t]
  \caption{Adaptation on meta-test tasks}
   \label{alg:mml-aux-test}
   \begin{flushleft}
    \textbf{Input}: A set of meta-test task datasets $\gD=\{\gD_1,\dots,\gD_M\}$ where $\gD_m=(\gD^{tr}_m,\gD^{te}_m)$.
    % $\gD_m=\{(u^m_i(t-T),\cdots,u^m_i(t-1),\ve^m_{ij}, y^{m}_i) : i\in\sV_b, (i,j)\in\mathbb{E}_b\}$ where $y^{m}_i$ is the next signal $u^m_i(t)$ for the $i$-th node. 
    Meta-initialized SDM ($\Phi$).
    Learning rate $\alpha$.\\
   \textbf{Output}: Adapted TDM $\theta'_m$ on the $m$-th task.
   \end{flushleft}
   \begin{algorithmic}[1]
   \STATE Initialize temporal modules $(\theta_1,\dots,\theta_M)$
%   \WHILE{not converged}
%   \STATE Sample a task $\gT_i$ from $\gT$
   \FOR{$\gD_m$ in $\gD$}
%   \STATE $\Phi'_m=\Phi - \alpha\nabla_{\Phi}\mathcal{L}(\gD^{tr}_m;\Phi,\theta_m)$
   \STATE $\theta'_m=\theta_m - \alpha\nabla_{\theta_m}\mathcal{L}(\gD^{tr}_m;\Phi,\theta_m)$
   \ENDFOR
%   \STATE $\Phi\leftarrow\Phi-\beta\nabla_{\Phi}\sum_{b=1}^{B}\sum_{k=1}^K\mathcal{L}^{aux}_{k}(\gD^{te}_b;\phi'_{b,k})$
%   \ENDWHILE
\end{algorithmic}
\end{algorithm}

%% file: sdm-generalization.tex
\begin{table}[t]
\centering
\small
% \scriptsize
\begin{tabular}{@{}ccc@{}}
\toprule
& Meta-train & Meta-test \\ \midrule
\# nodes ($N$) & \{256, 625\} & \{450, 800\} \\
\# edges per a node ($E$) & \{4, 8\} & \{3, 6, 10\}      \\
Initial frequency ($F$) & \{2, 5\} & \{3, 7\} \\ \bottomrule
\end{tabular}
\caption{Parameters for synthetic dataset.} \label{tab:syn-data}
\vspace{-1em}
\end{table}

\section{Spatial Derivative Modules: Reusable Modules}\label{sec:sdm-reuse}

We have claimed that the spatial modules provide reusable features associated with spatial derivatives such as $\nabla_x u,\nabla_y u$, and $\nabla^2_{x} u$ across different dynamics or PDEs.
While it has been shown that the data-driven approximation of spatial derivatives is more precise than that of finite difference method~\cite{Seo*2020Physics-aware,bar2019learning}, it is not guaranteed that the modules effectively provide transferrable parameters for different spatial resolution, discretization, and fluctuation of function values.
We explore whether the proposed spatial derivative modules based on graph networks can be used as a feature provider for different spatial functions and discretization.

\begin{table*}[t]
\centering
\footnotesize
% \scriptsize
\begin{tabular}{@{}ccccccc@{}}
\toprule
$(N,E,F)$ & (450,3,3) & (450,3,7) & (450,6,3) & (450,6,7) & (450,10,3) & (450,10,7) \\ \midrule
SDM (from scratch) & \begin{tabular}{@{}c@{}}1.337$\pm$0.044 \\ 7.278$\pm$0.225\end{tabular} & \begin{tabular}{@{}c@{}}7.111$\pm$0.148 \\ 51.544$\pm$0.148\end{tabular} & \begin{tabular}{@{}c@{}}1.152$\pm$0.043 \\ 5.997$\pm$0.083\end{tabular} & \begin{tabular}{@{}c@{}}7.206$\pm$0.180 \\ 47.527$\pm$0.768\end{tabular} & \begin{tabular}{@{}c@{}}1.112$\pm$0.036 \\ 5.353$\pm$0.193\end{tabular} & \begin{tabular}{@{}c@{}}7.529$\pm$0.241 \\ 47.356$\pm$0.560\end{tabular}\\ \cmidrule{2-7}
SDM (pretrained) & \begin{tabular}{@{}c@{}}\textbf{1.075$\pm$0.005} \\ \textbf{6.482$\pm$0.207}\end{tabular} & \begin{tabular}{@{}c@{}}\textbf{5.528$\pm$0.010} \\ \textbf{46.254$\pm$0.262}\end{tabular} & \begin{tabular}{@{}c@{}}\textbf{0.836$\pm$0.002} \\ \textbf{5.251$\pm$0.245}\end{tabular} & \begin{tabular}{@{}c@{}}\textbf{5.354$\pm$0.001} \\ \textbf{42.243$\pm$0.420}\end{tabular} & \begin{tabular}{@{}c@{}}\textbf{0.782$\pm$0.006} \\ \textbf{4.728$\pm$0.244}\end{tabular} & \begin{tabular}{@{}c@{}}\textbf{5.550$\pm$0.012} \\ \textbf{42.754$\pm$0.442}\end{tabular} \\ \midrule
$(N,E,F)$ & (800,3,3) & (800,3,7) & (800,6,3) & (800,6,7) & (800,10,3) & (800,10,7)
\\ \midrule
SDM (from scratch) & \begin{tabular}{@{}c@{}}1.022$\pm$0.030 \\ 7.196$\pm$0.159\end{tabular} & \begin{tabular}{@{}c@{}}5.699$\pm$0.242 \\ 49.602$\pm$0.715\end{tabular} & \begin{tabular}{@{}c@{}}0.789$\pm$0.021 \\ 5.386$\pm$0.136\end{tabular} & \begin{tabular}{@{}c@{}}5.179$\pm$0.069 \\ 42.509$\pm$1.080\end{tabular} & \begin{tabular}{@{}c@{}}0.718$\pm$0.010 \\ 4.536$\pm$0.204\end{tabular} & \begin{tabular}{@{}c@{}}5.517$\pm$0.110 \\ 39.642$\pm$1.173\end{tabular} \\ \cmidrule{2-7}
SDM (pretrained) & \begin{tabular}{@{}c@{}}\textbf{0.927$\pm$0.006} \\ \textbf{6.553$\pm$0.193}\end{tabular} & \begin{tabular}{@{}c@{}}\textbf{4.415$\pm$0.011} \\ \textbf{44.591$\pm$0.002}\end{tabular} & \begin{tabular}{@{}c@{}}\textbf{0.656$\pm$0.008} \\ \textbf{4.960$\pm$0.266}\end{tabular} & \begin{tabular}{@{}c@{}}\textbf{3.977$\pm$0.025} \\ \textbf{37.629$\pm$0.760}\end{tabular} & \begin{tabular}{@{}c@{}}\textbf{0.570$\pm$0.006} \\ \textbf{4.213$\pm$0.275}\end{tabular} & \begin{tabular}{@{}c@{}}\textbf{4.107$\pm$0.019} \\ \textbf{35.849$\pm$0.947}\end{tabular} \\ \bottomrule
\end{tabular}
\caption{Prediction error (MAE) of the first (top) and second (bottom) order spatial derivatives.}
\label{tab:syn-sdm-result}
\vspace{-1em}
\end{table*}

We perform two sets of experiments: evaluate few-shot learning performance (1) when SDM is trained from scratch; (2) when SDM is meta-initialized.
Fig.~\ref{fig:aux} shows how the graph signal and its discretization is changed over the different settings.
If the number of nodes is large, it can provide spatially high-resolution and thus, the spatial derivatives can be more precisely approximated.
Table~\ref{tab:syn-data} shows the parameters we used to generate synthetic datasets.
Note that meta-test data is designed to evaluate interpolation/extrapolation properties.
Initial frequency decides the degree of fluctuation (In Fig.~\ref{fig:aux}, the right one has higher $F$ than that of the left one.).
For each parameter combination, we generate 100 different snapshots based on a form in~\cite{long2018pde}:
\begin{equation}
\resizebox{.91\linewidth}{!}{$
    \displaystyle
    u_i =\sum_{|k|,|l| \leq F} \lambda_{k, l}\cos (k x_i+l y_i)+\gamma_{k, l}\sin (k x_i+l y_i),
$}
\end{equation}%
where $\lambda_{k,l},\gamma_{k,l}\sim\mathcal{N}\left(0,0.02\right)$. 
The index $i$ denotes the $i$-th node whose coordinate is $(x_i,y_i)$ in the 2D space $([0,2\pi]\times[0,2\pi])$ and $k, l$ are randomly sampled integers. 
From the synthetic data, the first- and second-order derivatives are analytically given and SDM is trained to approximate them.

The prediction results for spatial derivatives are shown in Table~\ref{tab:syn-sdm-result}.
The results show that the proposed module (SDM) is efficiently adaptable to different configuration on few samples from meta-initialized parameters compared to learning from scratch.
The finding implies that the parameters for spatial derivatives can be generally applicable across different spatial resolution, discretization, and function fluctuation.

%% file: experiment.tex
\section{Experimental Evaluation}
\subsection{Preliminary: Which Synthetic Dynamics Need to Be Generated?}
While Table~\ref{tab:syn-sdm-result} demonstrates that the PDE-independent representations are reusable across different configurations, it is still an open question: which topological configuration needs to be used to construct the synthetic dynamics?
According to Table~\ref{tab:syn-sdm-result}, the most important factor affecting error is an initial frequency ($F$), which determines min/max scales and fluctuation of function values, and it implies that the synthetic dynamics should be similarly scaled to a target dynamics.
We use the same topological configuration in Table~\ref{tab:syn-data} to generate synthetic datasets for a task in Sec.~\ref{sec:multistep-generation} and  Sec.~\ref{sec:graph-signal-regression}.
% We describe more details in Appendix.

\subsection{Multi-step Graph Signal Generation} \label{sec:multistep-generation}
\paragraph{Task.} We adopt a set of multi-step spatiotemporal sequence generation tasks to evaluate our proposed framework.
In each task, the data is a sequence of $L$ frames, where each frame is a set of observations on $N$ nodes in space.
Then, we train an auto-regressive model with the first $T$ frames ($T$-shot) and generate the following $L-T$ frames repeatedly from a given initial input ($T$-th frame) to evaluate its performance.
% We evaluate the performance of the model via the average of mean squared errors (MSEs) over all tasks. 

% \begin{figure*}[t]
%     \centering
%     \includegraphics[width=\textwidth]{fig/extreme_weather_nodes.pdf}
%     \caption{Visualization of the first 5 frames of one extended sequence in the extreme weather dataset. Dots represent the sampled points. Greenish (purplish) area is higher (lower) surface temperature.}
%     \label{fig:extremeweather}
% \end{figure*}

\paragraph{Datasets.} For all experiments, we generate meta-train tasks with the parameters described in Table~\ref{tab:syn-data} and  the target observations are 2 real-world datasets: (1) \textbf{AQI-CO}: national air quality index (AQI) observations~\cite{aqi2017}; (2) \textbf{ExtremeWeather (EW)}: the extreme weather dataset~\cite{racah2017extremeweather}.
For the AQI-CO dataset, we construct 12 meta-test tasks with the carbon monoxide (CO) ppm records from the first week of each month in 2015 at land-based stations. 
For the extreme weather dataset, we select the top-10 extreme weather events with the longest lasting time from the year 1984 and construct a meta-test task from each event with the observed surface temperatures at randomly sampled locations.
% \sscomment{about sampling}
Since each event lasts fewer than 20 frames, each task has a very limited amount of available data. 
In both datasets, graph signals are univariate.
Note that all quantities have fluidic properties such as diffusive and convection.
% Fig.~\ref{fig:extremeweather} shows the spatiotemporal dynamics of the extreme weather observations and sampled points.
More details are in the supplementary material.

\begin{table}[t]
\centering
\small
\begin{tabular}{@{}cccc@{}}
\toprule
                $T$-shot  & Method & AQI-CO & EW \\ \midrule
%  \multirow{2}{*}{3-shot} & PA-DGN (scratch) & 0.6762$\pm$0.1547 & 1.5477$\pm$0.5140 \\
    % & {\ours} (meta-init) & \textbf{0.5254$\pm$0.1272} & \textbf{1.3584$\pm$0.5749} \\ \midrule
  \multirow{3}{*}{5-shot} & FDM+RGN (scratch) & 0.029$\pm$0.004 & 0.988$\pm$0.557 \\
  & PA-DGN (scratch) & 0.036$\pm$0.009 & 0.965$\pm$0.138 \\
    & {\ours} (meta-init) & \textbf{0.025$\pm$0.006} & \textbf{0.917$\pm$0.075}  \\ \midrule
    \multirow{3}{*}{7-shot} & FDM+RGN (scratch) & 0.026$\pm$0.002 & 0.763$\pm$0.060\\
    & PA-DGN (scratch) & 0.023$\pm$0.002 & 0.748$\pm$0.020 \\ 
    & {\ours} (meta-init) & \textbf{0.018$\pm$0.002} & \textbf{0.727$\pm$0.009} \\ \midrule
    \multirow{3}{*}{10-shot} & FDM+RGN (scratch) & 0.021$\pm$0.001 & 0.709$\pm$0.003 \\
    & PA-DGN (scratch) & 0.015$\pm$0.001 & 0.416$\pm$0.015 \\
    & {\ours} (meta-init) & \textbf{0.012$\pm$0.001} & \textbf{0.407$\pm$0.025} \\
\bottomrule
\end{tabular}
\caption{Multi-step prediction results (MSE) and standard deviations on the two real-world datasets.}
\label{tab:real-world-exp}
\vspace{-1em}
\end{table}

\paragraph{Baselines.} We evaluate the performance of a physics-aware architecture (\textbf{PA-DGN}) \cite{Seo*2020Physics-aware}, which also consists of spatial derivative modules and recurrent graph networks (RGN), to see how the additional spatial information affects prediction performance for same architecture.
% We evaluate the performance of (1) a gated recurrent unit (\textbf{GRU}) and (2) a physics-aware architecture (\textbf{PA-DGN}) \citep{Seo*2020Physics-aware} which also consists of spatial derivative modules and recurrent graph networks (RGN).
Note that PA-DGN has same modules in {\ours} and the difference is that {\ours} utilizes meta-initialized spatial modules and PA-DGN is randomly initialized for learning from scratch on meta-test tasks.
Additionally, the spatial modules in PA-DGN is replaced by finite difference method (\textbf{FDM+RGN}) to see if the numerical method provides better PDE-agnostic representations.
The baselines and {\ours} are trained on the meta-test support set only to demonstrate how the additional spatial information is beneficial for few-shot learning tasks.
% (1) a Recurrent Graph Neural Network architecture (\textbf{RGN}) \cite{battaglia2018relational} and (2) a physics-aware architecture (\textbf{PA-DGN}) \cite{Seo*2020Physics-aware} with spatial derivative modules (SDM) and temporal derivative modules (TDM), where the SDM is a Graph Neural Network and the TDM is a Recurrent Graph Neural Network. For the RGN architecture, we evaluate its performance via (1) training on meta-test tasks only (\textbf{train from scratch}), (2) training on meta-train tasks and using the trained model weights for the initialization of models trained on meta-test tasks (\textbf{weight init}), (3) model-agnostic meta training (\textbf{MAML}) \cite{finn2017model}. For the PA-DGN architecture, in addition to train from scratch and weight init, we consider 2 versions of our proposed physics-aware modular meta-learning methods: (1) the vanilla modular meta-learning with auxiliary tasks in Algorithm \ref{alg:mml-aux-train} (\textbf{{\ours}-modular}) and (2) the MAML variation in Eq.~\ref{eq:maml1} and~\ref{eq:maml2} (\textbf{{\ours}-MAML}). Details of architectures and implementation of models are in the supplementary material.

\begin{table*}[t]
\centering
% \scriptsize
\small
\begin{tabular}{@{}ccccccc@{}}
\toprule
$T$-shot (Region) & GCN & GAT & GraphSAGE & GN & PA-DGN & {\ours} \\ \midrule
5-shot (USA) & 2.742$\pm$0.120 & 2.549$\pm$0.115 & 2.128$\pm$0.146 & 2.252$\pm$0.131 & 1.950$\pm$0.152 & \textbf{1.794}$\pm$\textbf{0.130} \\
10-shot (USA) & 2.371$\pm$0.095 & 2.178$\pm$0.066 & 1.848$\pm$0.206 & 1.949$\pm$0.115 & 1.687$\pm$0.104 & \textbf{1.567}$\pm$\textbf{0.103} \\ \midrule
5-shot (EU) & 1.218$\pm$0.218 & 1.161$\pm$0.234 & 1.165$\pm$0.248 & 1.181$\pm$0.210 & 0.914$\pm$0.167 & \textbf{0.781}$\pm$\textbf{0.019} \\
10-shot (EU) & 1.186$\pm$0.076 & 1.142$\pm$0.070 & 1.044$\pm$0.210 & 1.116$\pm$0.147 & 0.831$\pm$0.058 & \textbf{0.773}$\pm$\textbf{0.014} \\
\bottomrule
\end{tabular}
\caption{Graph signal regression results (MSE, $10^{-3}$) and standard deviations on the two regions of weather stations.}
\label{tab:ghcn-exp}
\vspace{-1em}
\end{table*}

\paragraph{Discussion.} Table~\ref{tab:real-world-exp} shows the multi-step prediction performance of our proposed framework against the baselines on real-world datasets.
Overall, PA-DGN and {\ours} show similar trend such that the prediction error is decreased as longer series are available for few-shot adaptation.
% There are two important findings: first, since the task is prediction of spatiotemporal observations, PA-DGN and {\ours} outperform GRU which does not utilize any spatial dynamics.
There are two important findings: first, with the similar expressive power in terms of the number of learnable parameters, the meta-initialized spatial modules provide high quality representations which are easily adaptable across different spatiotemporal dynamics in the real-world.
This performance gap demonstrates that we can get a stronger inductive bias from synthetic datasets without knowing PDE-specific information.
Second, the contribution of the meta-initialization is more significant when the length of available sequence is shorter ($T=5$) and this demonstrates when the meta-initialization is particularly effective.
Finally, the finite difference method provides proxies of exact spatial derivatives and the representations are useful particularly when $T=5$ but its performance is rapidly saturated and it comes from the gap between the learnable spatial modules and fixed numerical coefficients.
The results provide a new point of view on how to utilize synthetic or simulated datasets to handle challenges caused by limited datasets.

% \begin{figure}[t]
%     \centering
%     \subfigure{\includegraphics[width=0.35\linewidth]{fig/usa-stations.pdf}}
%     \hfill
%     \subfigure{\includegraphics[width=0.26\linewidth]{fig/ghcn-usa-syn-xy.pdf}}
%     \hfill
%     \subfigure{\includegraphics[width=0.26\linewidth]{fig/ghcn-usa-syn-value.pdf}}
%     \caption{(left) GHCN weather stations in the USA, (center) spatial distribution of the GHCN stations and synthetic data points, and (right) graph signal scales from the two datasets.}
%     \label{fig:ghcn}
% \end{figure}

\subsection{Graph Signal Regression}\label{sec:graph-signal-regression}
\paragraph{Task, datasets, and baselines.} \cite{defferrard2019deepsphere} conducted a graph signal regression task: predict the temperature $x_t$ from the temperature on the previous 5 days ($x_{t-5}:x_{t-1}$).
We split the \textbf{GHCN} dataset spatially into two regions: (1) the USA (1,705 stations) and (2) Europe (EU) (703 stations) where there are many weather stations full functioning. 
In this task, the number of shots is defined as the number of input and output pairs to train a model.

As the input length is fixed, more variants of graph neural networks are considered as baselines. 
We concatenate the 5-step signals and feed it into Graph convolutional networks (\textbf{GCN})~\cite{kipf2016semi}, Graph attention networks (\textbf{GAT})~\cite{velickovic2018graph}, \textbf{GraphSAGE}~\cite{hamilton2017inductive}, and Graph networks (\textbf{GN})~\cite{sanchez2018graph} to predict next signals across all nodes.

\paragraph{Discussion.}
Table~\ref{tab:ghcn-exp} shows the results of the graph signal regression task across different baselines and the proposed method.
There are two patterns in the results. 
First, although in general we observe an improvement in performance for all methods when we move from the 5-shot setting to the 10-shot setting, {\ours}’s performance yields the smallest error. Second, for the EU dataset, while 5-shot seems enough to achieve stable performance, it demonstrates that the PDE-independent representations make the regression error converge to a lower level. 
Overall, the experimental results prove that the learned spatial representations from simulated dynamics are beneficial for learning on limited data.

%% file: related-work.tex
\section{Related Work}

\paragraph{Physics-informed learning.}
Since physics-informed neural networks are introduced in~\cite{raissi2019physics}, which find that a solution of a PDE can be discovered by neural networks, physical knowledge has been used as an inductive bias for deep neural networks.
Advection-diffusion equation is incorporated with deep neural networks for sea-surface temperature dynamics~\cite{de2018deep}.
\cite{lutter2018deep,greydanus2019hamiltonian} show that Lagrangian/Hamiltonian mechanics can be imposed to learn the equations of motion of a mechanical system and~\cite{seo2019differentiable} regularize a graph neural network with a specific physics equation.
Rather than using explicitly given equations, physics-inspired inductive bias is also used for reasoning dynamics of discrete objects~\cite{battaglia2016interaction} and continuous quantities~\cite{Seo*2020Physics-aware}.
\cite{long2018pde} propose a numeric-symbolic hybrid deep neural network designed to discover PDEs from observed dynamic data.
To the best of our knowledge, we are the first to provide a framework to use the physics-inspired inductive bias under the meta-learning settings to tackle the limited data issue which is common for real-world data such as extreme weather events~\cite{racah2017extremeweather}.

\paragraph{Meta-learning.}
% In classical machine learning paradigm, the aim is to learn a model to perform a specific task, i.e., the task specification is \emph{a priori} known. 
% Therefore, the model has to be learned from scratch as the task changes.
The aim of meta-learning is to enable learning parameters which can be used for new tasks unknown at the time of learning, leading to agile models which adapt to a new task utilizing only a few samples. %~\cite{schmidhuber1987,thrun1998learning}.
% As a result, the primary goal of meta-learning is task-agnostic learning.
Based on how the knowledge from the related tasks is used, meta-learning methods have been classified as optimization-based~\cite{ravi2016optimization,finn2017model}, model-based~\cite{munkhdalai2017meta,duan2017one,mishra2018a}, and metric-based~\cite{snell2017prototypical}.
% Optimization-based methods, aim to learn the optimization process by casting the algorithm design as a learning problem.
% Based on how the knowledge from the related tasks is used, meta-learning methods have been classified as a) optimization-based, b) model-based, and c) metric-based. Optimization-based methods, aim to learn the optimization process by casting the algorithm design as a learning problem \cite{andrychowicz2016learning,ravi2016optimization,duan2017one,finn2017model,nichol2018first,antoniou2018train,rusu2018meta,grant2018recasting}.
% In case of model-based approaches~\cite{santoro2016one,munkhdalai2017meta,mishra2018a}, the learned network weights are used as an initialization for a new task, while keeping the optimizer fixed \cite{yu2018one}.
% Further, modular meta-learning methods aim to identify and learn transferable modules or structural blocks which can be used in combination to for a new task \cite{alet2018modular,alet2019neural,chen2019modular}.
Recently, another branch of meta-learning has been introduced to more focus on finding a set of reusable modules as components of a solution to a new task.
\cite{alet2019neural} provide a framework, structured modular meta-learning, where a finite number of modules are introduced as task-independent modules and an optimal structure combining the modules is found from a limited number of data.
% \cite{chen2019modular} introduces techniques to automatically discover task-independent/dependent modules based on Bayesian shrinkage to find more adaptable modules.
To our knowledge, none of the above works provide a solution for modeling physics-related spatiotemporal dynamics where it is hard to generate enough tasks for meta-initialization.

%% file: conclusion.tex
\section{Conclusion}
In this paper, we propose a framework for physics-aware meta-learning with auxiliary tasks. 
By incorporating PDE-independent/-invariant knowledge from simulated data, the framework provide reusable features to meta-test tasks with a limited amount of data. 
Experiments show that auxiliary tasks and physics-aware meta-learning help construct reusable modules that improve the performance of spatiotemporal predictions when data is limited.
% Designing and identifying the most useful auxiliary tasks and data will be the focus of our future work.